%% file: main.tex
\documentclass{article}

\usepackage[nonatbib,final]{neurips_2021}

\usepackage[utf8]{inputenc} 
\usepackage[T1]{fontenc}    
\usepackage{hyperref}       
\usepackage{url}            
\usepackage{booktabs}       
\usepackage{amsfonts}       
\usepackage{nicefrac}       
\usepackage{microtype}      
\usepackage{graphicx}
\usepackage{amsmath}
\usepackage{color}
\usepackage[dvipsnames]{xcolor}
\usepackage{comment}
\usepackage{multirow}
\usepackage{pifont}
\usepackage{hyperref}
\usepackage{threeparttable}

\newcommand{\yes}{\ding{51}}

\input{math_commands}

\newcommand{\attn}[1]{\mathrm{Attn}\rbr{#1}}
\newcommand{\softmax}[1]{\mathrm{softmax}\rbr{#1}}

\title{Do Transformers Really Perform Bad \\ for Graph Representation? }

\author{%
Chengxuan Ying$^1$\thanks{Interns at MSRA.}, \ Tianle Cai$^2$,\ Shengjie Luo$^3$\footnotemark[1], \\ \textbf{Shuxin Zheng}$^4$\thanks{Corresponding authors.},\  \textbf{Guolin Ke}$^4$,\ \textbf{Di He}$^4$\footnotemark[2],\ \textbf{Yanming Shen}$^1$,\ \textbf{Tie-Yan Liu}$^4$\\
$^1$Dalian University of Technology \quad $^2$Princeton University \\
$^3$Peking University \quad $^4$Microsoft Research Asia\\
\texttt{\footnotesize yingchengsyuan@gmail.com, tianle.cai@princeton.edu, luosj@stu.pku.edu.cn} \\
\texttt{\footnotesize \{shuz\footnotemark[2], guoke, dihe\footnotemark[2], tyliu\}@microsoft.com, shen@dlut.edu.cn}}

\begin{document}

\maketitle

\begin{abstract}
The Transformer architecture has become a dominant choice in many domains, such as natural language processing and computer vision. Yet, it has not achieved competitive performance on popular leaderboards of graph-level prediction compared to mainstream GNN variants. Therefore, it remains a mystery how Transformers could perform well for graph representation learning. In this paper, we solve this mystery by presenting Graphormer, which is built upon the standard Transformer architecture, and could attain excellent results on a broad range of graph representation learning tasks, especially on the recent OGB Large-Scale Challenge. Our key insight to utilizing Transformer in the graph is the necessity of effectively encoding the structural information of a graph into the model. To this end, we propose several simple yet effective structural encoding methods to help Graphormer better model graph-structured data. Besides, we mathematically characterize the expressive power of Graphormer and exhibit that with our ways of encoding the structural information of graphs, many popular GNN variants could be covered as the special cases of Graphormer. The code and models of Graphormer will be made publicly available at \url{https://github.com/Microsoft/Graphormer}.

\end{abstract}

\section{Introduction}

The Transformer~\cite{vaswani2017attention} is well acknowledged as the most powerful neural network in modelling sequential data, such as natural language~\cite{devlin2019bert,liu2019roberta,brown2020language} and speech~\cite{gulati2020conformer}. Model variants built upon Transformer have also been shown great performance in computer vision~\cite{dosovitskiy2020image,liu2021Swin} and programming language~\cite{hellendoorn2019global,zugner2020language,peng2021could}. 
However, to the best of our knowledge, Transformer has still not been the de-facto standard on public graph representation leaderboards~\cite{hu2020open,dwivedi2020benchmarking,hu2021ogb}. There are many attempts of leveraging Transformer into the graph domain, but the only effective way is replacing some key modules (e.g., feature aggregation) in classic GNN variants by the softmax attention ~\cite{velivckovic2018graph, cai2020graph,hu2020heterogeneous,wang2020direct,zhang2020graph,rong2020self,dwivedi2021generalization}. 
Therefore, it is still an open question whether Transformer architecture is suitable to model graphs and how to make it work in graph representation learning.

In this paper, we give an affirmative answer by developing Graphormer, which is directly built upon the standard Transformer, and achieves state-of-the-art performance on a wide range of graph-level prediction tasks, including the very recent Open Graph Benchmark Large-Scale Challenge (OGB-LSC)~\cite{hu2021ogb}, and several popular leaderboards (e.g., OGB~\cite{hu2020open}, Benchmarking-GNN~\cite{dwivedi2020benchmarking}). The Transformer is originally designed for sequence modeling. To utilize its power in graphs, we believe the key is to properly incorporate structural information of graphs into the model. Note that for each node $i$, the self-attention only calculates the semantic similarity between $i$ and other nodes, without considering the structural information of a graph reflected on the nodes and the relation between node pairs. Graphormer incorporates several effective structural encoding methods to leverage such information, which are described below.

First, we propose a \emph{Centrality Encoding} in Graphormer to capture the node importance in the graph. In a graph, different nodes may have different importance, e.g., celebrities are considered to be more influential than the majority of web users in a social network. However, such information isn't reflected in the self-attention module as it calculates the similarities mainly using the node semantic features. To address the problem, we propose to encode the node centrality in Graphormer. In particular, we leverage the \emph{degree centrality} for the centrality encoding, where a learnable vector is assigned to each node according to its degree and added to the node features in the input layer. Empirical studies show that simple centrality encoding is effective for Transformer in modeling the graph data.

Second, we propose a novel \emph{Spatial Encoding} in Graphormer to capture the structural relation between nodes. One notable geometrical property that distinguishes graph-structured data from other structured data, e.g., language, images, is that there does not exist a canonical grid to embed the graph. In fact, nodes can only lie in a non-Euclidean space and are linked by edges. To model such structural information, for each node pair, we assign a learnable embedding based on their spatial relation. Multiple measurements in the literature could be leveraged for modeling spatial relations. For a general purpose, we use the distance of the shortest path between any two nodes as a demonstration, which will be encoded as a bias term in the softmax attention and help the model accurately capture the spatial dependency in a graph. In addition, sometimes there is additional spatial information contained in edge features, such as the type of bond between two atoms in a molecular graph. We design a new edge encoding method to further take such signal into the Transformer layers. To be concrete, for each node pair, we compute an average of dot-products of the edge features and learnable embeddings along the shortest path, then use it in the attention module. Equipped with these encodings, Graphormer could better model the relationship for node pairs and represent the graph.

By using the proposed encodings above, we further mathematically show that Graphormer has strong expressiveness as many popular GNN variants are just its special cases. The great capacity of the model leads to state-of-the-art performance on a wide range of tasks in practice. On the large-scale quantum chemistry regression dataset\footnote{\url{https://ogb.stanford.edu/kddcup2021/pcqm4m/}} in the very recent Open Graph Benchmark Large-Scale Challenge (OGB-LSC)~\cite{hu2021ogb}, Graphormer outperforms most mainstream GNN variants by more than 10\% points in terms of the relative error. On other popular leaderboards of graph representation learning (e.g., MolHIV, MolPCBA, ZINC)~\cite{hu2020open,dwivedi2020benchmarking}, Graphormer also surpasses the previous best results, demonstrating the potential and adaptability of the Transformer architecture.


\section{Preliminary}
In this section, we recap the preliminaries in Graph Neural Networks and Transformer.

\paragraph{Graph Neural Network (GNN).} 
Let $G = \rbr{V, E}$ denote a graph where $V = \cbr{v_1, v_2, \cdots, v_n}$, $n = \abs{V}$ is the number of nodes. Let the feature vector of node $v_i$ be $x_i$. GNNs aim to learn representation of nodes and graphs. Typically, modern GNNs follow a learning schema that iteratively updates the representation of a node by aggregating representations of its first or higher-order neighbors. We denote $h^{(l)}_i$ as the representation of $v_i$ at the $l$-th layer and define $h_i^{(0)} = x_i$. The $l$-th iteration of aggregation could be characterized by AGGREGATE-COMBINE step as
\begin{equation}
a_{i}^{(l)}=\text { AGGREGATE }^{(l)}\left(\left\{h_{j}^{(l-1)}: j \in \mathcal{N}(v_i)\right\}\right), \quad
h_{i}^{(l)}=\text { COMBINE }^{(l)}\left(h_{i}^{(l-1)}, a_{i}^{(l)}\right) ,
\label{eqn:agg-comb}
\end{equation}
where $\mathcal{N}(v_i)$ is the set of first or higher-order neighbors of $v_i$. The AGGREGATE function is used to gather the information from neighbors. Common aggregation functions include MEAN, MAX, SUM, which are used in different architectures of GNNs~\cite{kipf2016semi,hamilton2017inductive,velivckovic2018graph,xu2018how}. The goal of COMBINE function is to fuse the information from neighbors into the node representation. 

In addition, for graph representation tasks, a READOUT function is designed to aggregate node features $h_i^{(L)}$ of the final iteration into the representation $h_G$ of the entire graph $G$:
\begin{equation}
h_{G}=\operatorname{READOUT}\left(\left\{h_{i}^{(L)} \mid v_i \in G \right\}\right).
\end{equation}
READOUT can be implemented by a simple permutation invariant function such as summation~\cite{xu2018how} or a more sophisticated graph-level pooling function~\cite{baek2021accurate}.

\paragraph{Transformer.}
The Transformer architecture consists of a composition of Transformer layers~\cite{vaswani2017attention}. Each Transformer layer has two parts: a self-attention module and a position-wise feed-forward network (FFN). Let $H = \sbr{h_1^\top, \cdots, h_n^\top}^\top\in\Rbb^{n\times d}$ denote the input of self-attention module where $d$ is the hidden dimension and $h_i\in\Rbb^{1\times d}$ is the hidden representation at position $i$. The input $H$ is projected by three matrices $W_Q\in\Rbb^{d\times d_K}, W_K\in\Rbb^{d\times d_K}$ and $ W_V\in\Rbb^{d\times d_V}$ to the corresponding representations $Q, K, V$. The self-attention is then calculated as:
\begin{align}
    Q = HW_Q,\quad K = HW_K,\quad V = HW_V,\label{eqn:attention-alpha}\\
    A = \frac{QK^\top}{\sqrt{d_K}}, \quad \attn{H} = \softmax{A}V,\label{eqn:attention-alpha2}
\end{align}
where $A$ is a matrix capturing the similarity between queries and keys. For simplicity of illustration, we consider the single-head self-attention and assume $d_K = d_V = d$. The extension to the multi-head attention is standard and straightforward, and we omit bias terms for simplicity.


\section{Graphormer}

In this section, we present our Graphormer for graph tasks. First, we elaborate on several key designs in the Graphormer, which serve as an inductive bias in the neural network to learn the graph representation. We further provide the detailed implementations of Graphormer. Finally, we show that our proposed Graphormer is more powerful since popular GNN models~\cite{kipf2016semi,xu2018how,hamilton2017inductive} are its special cases.

\subsection{Structural Encodings in Graphormer}
\label{sec:incorporate_graph_structure}

\begin{figure}[t]
    \centering
    \includegraphics[width=0.8\textwidth]{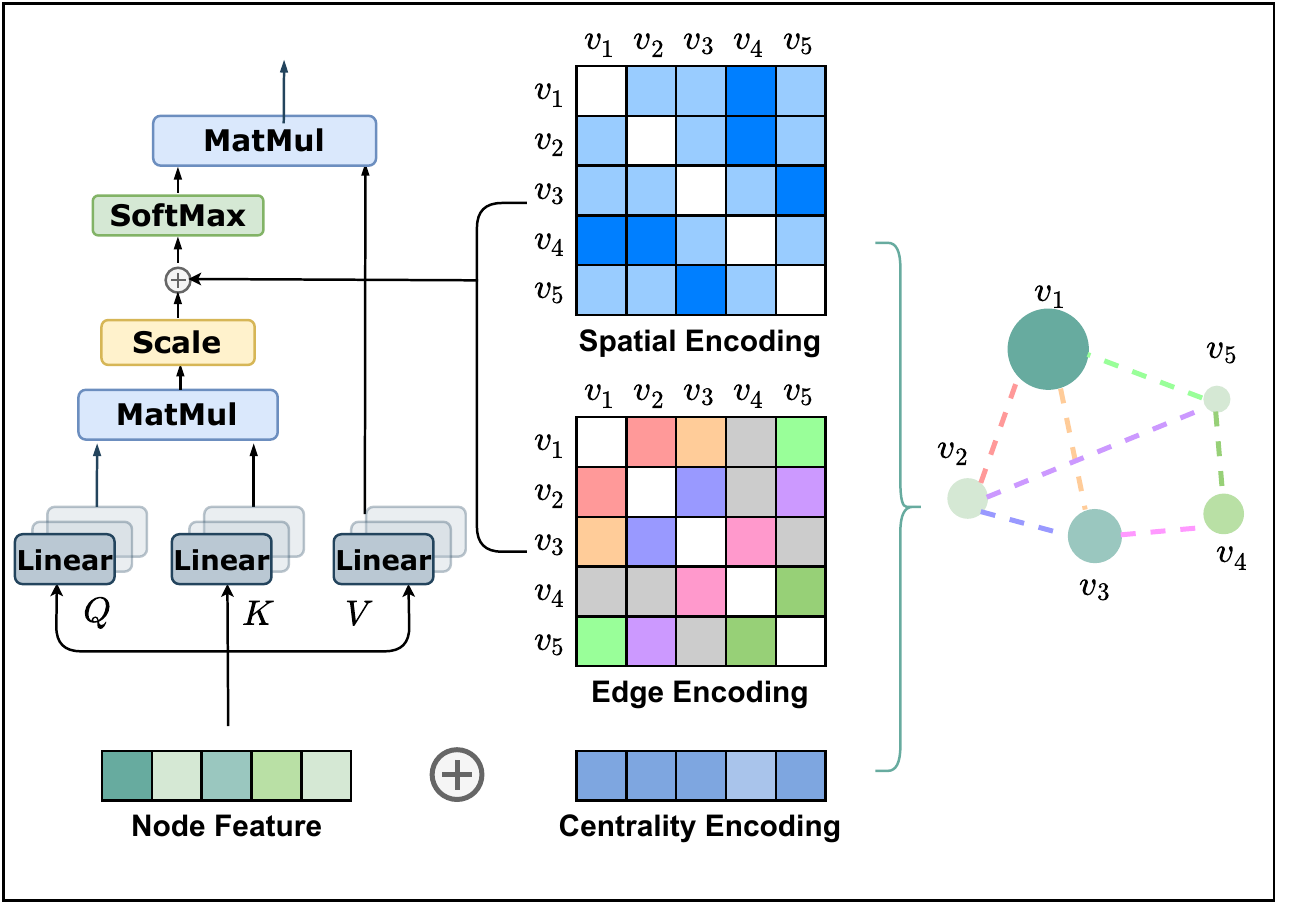}
    \caption{An illustration of our proposed centrality encoding, spatial encoding, and edge encoding in Graphormer.}
    \label{fig:model}
\end{figure}

As discussed in the introduction, it is important to develop ways to leverage the structural information of graphs into the Transformer model. To this end, we present three simple but effective designs of encoding in Graphormer. See Figure \ref{fig:model} for an illustration.

\subsubsection{Centrality Encoding}
In Eq.\ref{eqn:attention-alpha2}, the attention distribution is calculated based on the semantic correlation between nodes. However, node centrality, which measures how important a node is in the graph, is usually a strong signal for graph understanding. For example, celebrities who have a huge number of followers are important factors in predicting the trend of a social network~\cite{marwick2011see,marshall2010promotion}. 
Such information is neglected in the current attention calculation, and we believe it should be a valuable signal for Transformer models. 

In Graphormer, we use the degree centrality, which is one of the standard centrality measures in literature, as an additional signal to the neural network. To be specific, we develop a \emph{Centrality Encoding} which assigns each node two real-valued embedding vectors according to its indegree and outdegree. As the centrality encoding is applied to each node, we simply add it to the node features as the input.
\begin{align}
\label{eqn:degree_encoding}
    h_i^{(0)} = x_i + z^-_{\text{deg}^{-}(v_i)} + z^+_{\text{deg}^{+}(v_i)},
\end{align}
where $z^{-}, z^{+} \in \mathbb{R}^d$ are learnable embedding vectors specified by the indegree $\text{deg}^{-}(v_i)$ and outdegree $\text{deg}^{+}(v_i)$ respectively. For undirected graphs, $\text{deg}^{-}(v_i)$ and $\text{deg}^{+}(v_i)$ could be unified to $\text{deg}(v_i)$. By using the centrality encoding in the input, the softmax attention can catch the node importance signal in the queries and the keys. Therefore the model can capture both the semantic correlation and the node importance in the attention mechanism.

\subsubsection{Spatial Encoding} 
An advantage of Transformer is its global receptive field. In each Transformer layer, each token can attend to the information at any position and then process its representation. But this operation has a byproduct problem that the model has to explicitly specify different positions or encode the positional dependency (such as locality) in the layers. For sequential data, one can either give each position an embedding (i.e., absolute positional encoding \cite{vaswani2017attention}) as the input or encode the relative distance of any two positions (i.e., relative positional encoding \cite{raffel2019exploring,shaw2018self}) in the Transformer layer. 

However, for graphs, nodes are not arranged as a sequence. They can lie in a multi-dimensional spatial space and are linked by edges. To encode the structural information of a graph in the model, we propose a novel \emph{Spatial Encoding}. Concretely, for any graph $G$, we consider a function $\phi\rbr{v_i, v_j}:V\times V\to\Rbb$ which measures the spatial relation between $v_i$ and $v_j$ in graph $G$. The function $\phi$ can be defined by the connectivity between the nodes in the graph. In this paper, we choose $\phi(v_i,v_j)$ to be the distance of the shortest path (SPD) between $v_i$ and $v_j$ if the two nodes are connected. If not, we set the output of $\phi$ to be a special value, i.e., -1. We assign each (feasible) output value a learnable scalar which will serve as a bias term in the self-attention module. Denote $A_{ij}$ as the $(i,j)$-element of the Query-Key product matrix $A$, we have:
\begin{align}
\label{eqn:rnpe}
    A_{ij}=\frac{(h_iW_{Q})(h_jW_{K})^T}{\sqrt{d}}  + b_{\phi(v_i,v_j)},
\end{align}
where $b_{\phi(v_i,v_j)}$ is a learnable scalar indexed by $\phi(v_i,v_j)$, and shared across all layers.

Here we discuss several benefits of our proposed method. First,  compared to conventional GNNs described in Section 2, where the receptive field is restricted to the neighbors, we can see that in Eq.~\eqref{eqn:rnpe}, the Transformer layer provides a global information that each node can attend to all other nodes in the graph. Second, by using $b_{\phi(v_i,v_j)}$, each node in a single Transformer layer can adaptively attend to all other nodes according to the graph structural information. For example, if $b_{\phi(v_i,v_j)}$ is learned to be a decreasing function with respect to $\phi(v_i,v_j)$,  for each node, the model will likely pay more attention to the nodes near it and pay less attention to the nodes far away from it.

\subsubsection{Edge Encoding in the Attention}\label{sec:edge}
In many graph tasks, edges also have structural features, e.g., in a molecular graph, atom pairs may have features describing the type of bond between them. Such features are important to the graph representation, and encoding them together with node features into the network is essential. There are mainly two edge encoding methods used in previous works. In the first method, the edge features are added to the associated nodes' features~\cite{hu2020open,li2020deepergcn}. In the second method, for each node, its associated edges' features will be used together with the node features in the aggregation~\cite{gilmer2017neural,xu2018how,kipf2016semi}. However, such ways of using edge feature only propagate the edge information to its associated nodes, which may not be an effective way to leverage edge information in representation of the whole graph.

To better encode edge features into attention layers, we propose a new edge encoding method in Graphormer. The attention mechanism needs to estimate correlations for each node pair $(v_i, v_j)$, and we believe the edges connecting them should be considered in the correlation as in \cite{lin2018multi, wang2020direct}. For each ordered node pair $(v_i, v_j)$, we find (one of) the shortest path $\text{SP}_{ij}=(e_1,e_2,...,e_N)$ from $v_i$ to $v_j$, and compute an average of the dot-products of the edge feature and a learnable embedding along the path. The proposed edge encoding incorporates edge features via a bias term to the attention module. Concretely, we modify the $(i,j)$-element of $A$ in Eq.~\eqref{eqn:attention-alpha} further with the edge encoding $c_{ij}$ as:

\begin{align}
    A_{ij}=\frac{(h_iW_{Q})(h_jW_{K})^T}{\sqrt{d}} + b_{\phi(v_i,v_j)} + c_{ij},\ \text{where}\ c_{ij}=\frac{1}{N}\sum_{n=1}^{N} x_{e_n}(w^{E}_{n})^T,
    \label{eqn:attn-edge}
\end{align}

where $x_{e_n}$ is the feature of the $n$-th edge $e_n$ in $\text{SP}_{ij}$, $w_n^{E}\in \mathbb{R}^{d_E}$ is the $n$-th weight embedding, and $d_E$ is the dimensionality of edge feature. 

\subsection{Implementation Details of Graphormer}\label{sec:model}

\paragraph{Graphormer Layer.} Graphormer is built upon the original implementation of classic Transformer encoder described in ~\cite{vaswani2017attention}. In addition, we apply the layer normalization (LN) before the multi-head self-attention (MHA) and the feed-forward blocks (FFN) instead of after~\cite{xiong2020layer}. This modification has been unanimously adopted by all current Transformer implementations because it leads to more effective optimization~\cite{narang2021transformer}. Especially, for FFN sub-layer, we set the dimensionality of input, output, and the inner-layer to the same dimension with $d$. We formally characterize the Graphormer layer as below:

\begin{align}
    h^{'(l)} &= \text{MHA}(\text{LN}(h^{(l-1)})) + h^{(l-1)}\\
    h^{(l)} &= \text{FFN}(\text{LN}(h^{'(l)})) + h^{'(l)}
\end{align}

\paragraph{Special Node.} As stated in the previous section, various graph pooling functions are proposed to represent the graph embedding. Inspired by ~\cite{gilmer2017neural}, in Graphormer, we add a special node called {\tt[VNode]} to the graph, and make connection between {\tt[VNode]} and each node individually. In the AGGREGATE-COMBINE step, the representation of {\tt[VNode]} has been updated as normal nodes in graph, and the representation of the entire graph $h_G$ would be the node feature of {\tt[VNode]} in the final layer. In the BERT model~\cite{devlin2019bert,liu2019roberta}, there is a similar token, i.e., {\tt[CLS]}, which is a special token attached at the beginning of each sequence, to represent the sequence-level feature on downstream tasks. While the {\tt[VNode]} is connected to all other nodes in graph, which means the distance of the shortest path is $1$ for any $\phi({\tt[VNode]}, v_j)$ and $\phi(v_i, {\tt[VNode]})$, the connection is not physical. To distinguish the connection of physical and virtual, inspired by ~\cite{ke2020rethinking}, we reset all spatial encodings for $b_{\phi({\tt[VNode]}, v_j)}$ and $b_{\phi(v_i, {\tt[VNode]})}$ to a distinct learnable scalar.

\subsection{How Powerful is Graphormer?}
\label{sec:expressive}
In the previous subsections, we introduce three structural encodings and the architecture of Graphormer. Then a natural question is: \emph{Do these modifications make Graphormer more powerful than other GNN variants? } In this subsection, we first give an affirmative answer by showing that Graphormer can represent the AGGREGATE and COMBINE steps in popular GNN models:

\begin{fact}
\label{fact:recover_gnn}
By choosing proper weights and distance function $\phi$, the Graphormer layer can represent AGGREGATE and COMBINE steps of popular GNN models such as GIN, GCN, GraphSAGE.
\end{fact}
The proof sketch to derive this result is: 1) Spatial encoding enables self-attention module to distinguish neighbor set $\Ncal\rbr{v_i}$ of node $v_i$ so that the softmax function can calculate mean statistics over $\Ncal\rbr{v_i}$; 2) Knowing the degree of a node, mean over neighbors can be translated to sum over neighbors; 3) With multiple heads and FFN, representations of $v_i$ and $\Ncal\rbr{v_i}$ can be processed separately and combined together later. We defer the proof of this fact to Appendix A.

Moreover, we show further that by using our spatial encoding, Graphormer can go beyond classic message passing GNNs whose expressive power is no more than the 1-Weisfeiler-Lehman (WL) test. We give a concrete example in Appendix A to show how Graphormer helps distinguish graphs that the 1-WL test fails to.

\paragraph{Connection between Self-attention and Virtual Node.} 
Besides the superior expressiveness than popular GNNs, we also find an interesting connection between using self-attention and the virtual node heuristic~\cite{gilmer2017neural,li2017learning,ishiguro2019graph,hu2020open}. As shown in the leaderboard of OGB~\cite{hu2020open}, the virtual node trick, which augments graphs with additional supernodes that are connected to all nodes in the original graphs, can significantly improve the performance of existing GNNs. Conceptually, the benefit of the virtual node is that it can aggregate the information of the \emph{whole graph} (like the READOUT function) and then propagate it to \emph{each node}. However, a naive addition of a supernode to a graph can potentially lead to inadvertent over-smoothing of information propagation~\cite{ishiguro2019graph}. We instead find that such a graph-level aggregation and propagation operation can be naturally fulfilled by vanilla self-attention without additional encodings. Concretely, we can prove the following fact:
\begin{fact}
\label{fact:readout}
By choosing proper weights, every node representation of the output of a Graphormer layer without additional encodings can represent MEAN READOUT functions.
\end{fact}
This fact takes the advantage of self-attention that each node can attend to all other nodes. Thus it can simulate graph-level READOUT operation to aggregate information from the whole graph. Besides the theoretical justification, we empirically find that Graphormer does not encounter the problem of over-smoothing, which makes the improvement scalable. The fact also inspires us to introduce a special node for graph readout (see the previous subsection).


\section{Experiments}

We first conduct experiments on the recent OGB-LSC~\cite{hu2021ogb} quantum chemistry regression (i.e., PCQM4M-LSC) challenge, which is currently the biggest graph-level prediction dataset and contains more than 3.8M graphs in total. Then, we report the results on the other three popular tasks: ogbg-molhiv, ogbg-molpcba and ZINC, which come from the OGB~\cite{hu2020open} and benchmarking-GNN~\cite{dwivedi2020benchmarking} leaderboards. Finally, we ablate the important design elements of Graphormer. A detailed description of datasets and training strategies could be found in Appendix B.

\subsection{OGB Large-Scale Challenge}
\paragraph{Baselines.} We benchmark the proposed Graphormer with GCN~\cite{kipf2016semi} and GIN~\cite{xu2018how}, and their variants with virtual node (-VN)~\cite{gilmer2017neural}. They achieve the state-of-the-art valid and test mean absolute error (MAE) on the official leaderboard\footnote{\url{https://github.com/snap-stanford/ogb/tree/master/examples/lsc/pcqm4m\#performance}}~\cite{hu2021ogb}. In addition, we compare to GIN's multi-hop variant~\cite{brossard2020graph}, and 12-layer deep graph network DeeperGCN~\cite{li2020deepergcn}, which also show promising performance on other leaderboards. We further compare our Graphormer with the recent Transformer-based graph model GT~\cite{dwivedi2021generalization}.

\paragraph{Settings.} We primarily report results on two model sizes: \textbf{Graphormer} ($L=12, d=768$), and a smaller one \textbf{Graphormer$_{\small \textsc{Small}}$\xspace} ($L=6, d=512$). Both the number of attention heads in the attention module and the dimensionality of edge features $d_E$ are set to 32. We use AdamW as the optimizer, and set the hyper-parameter $\epsilon$ to 1e-8 and $(\beta1,\beta2)$ to (0.99,0.999). The peak learning rate is set to 2e-4 (3e-4 for \textbf{Graphormer$_{\small \textsc{Small}}$\xspace}) with a 60k-step warm-up stage followed by a linear decay learning rate scheduler. The total training steps are 1M. The batch size is set to 1024. All models are trained on 8 NVIDIA V100 GPUS for about 2 days.

\begin{table}[ht]
\small
\centering
\caption{Results on PCQM4M-LSC. * \ indicates the results are cited from the official leaderboard~\cite{hu2021ogb}. }
\label{tab:pcq-table}
\begin{tabular}{c|c|cc}
\toprule
method            & \#param. & train MAE     & validate MAE       \\ \hline
GCN~\cite{kipf2016semi}  & 2.0M & 0.1318   & 0.1691 (0.1684*)  \\

GIN~\cite{xu2018how} & 3.8M & 0.1203 & 0.1537 (0.1536*)   \\ 

GCN-{\scriptsize VN} ~\cite{kipf2016semi,gilmer2017neural} & 4.9M & 0.1225  & 0.1485 (0.1510*)  \\

GIN-{\scriptsize VN}~\cite{xu2018how,gilmer2017neural} & 6.7M & 0.1150   & 0.1395 (0.1396*)   \\ 

GINE-{\scriptsize VN} ~\cite{brossard2020graph,gilmer2017neural} & 13.2M & 0.1248 & 0.1430  \\ 

DeeperGCN-{\scriptsize VN}~\cite{li2020deepergcn,gilmer2017neural} & 25.5M & 0.1059        & 0.1398   \\

\hline
GT~\cite{dwivedi2021generalization}  & 0.6M & 0.0944 & 0.1400  \\ 

GT-{\scriptsize Wide}~\cite{dwivedi2021generalization} & 83.2M  & 0.0955 & 0.1408 \\ 

\hline

Graphormer$_{\small \textsc{Small}}$ & 12.5M &  0.0778  & 0.1264   \\ 
Graphormer & 47.1M  & \textbf{0.0582} & \textbf{0.1234}    \\
\bottomrule
\end{tabular}
\end{table}

\paragraph{Results.} Table \ref{tab:pcq-table} summarizes performance comparisons on PCQM4M-LSC dataset. From the table, GIN-{\scriptsize VN} achieves the previous state-of-the-art validate MAE of 0.1395. The original implementation of GT~\cite{dwivedi2021generalization} employs a hidden dimension of 64 to reduce the total number of parameters. For a fair comparison, we also report the result by enlarging the hidden dimension to 768, denoted by GT-{\scriptsize Wide}, which leads to a total number of parameters of 83.2M. While, both GT and GT-{\scriptsize Wide} do not outperform GIN-{\scriptsize VN} and DeeperGCN-{\scriptsize VN}. Especially, we do not observe a performance gain along with the growth of parameters of GT.

Compared to the previous state-of-the-art GNN architecture, Graphormer noticeably surpasses GIN-{\scriptsize VN} by a large margin, e.g., 11.5\% relative validate MAE decline. By using the ensemble with ExpC~\cite{yang2020breaking}, we got a 0.1200 MAE on complete test set and won the first place of the graph-level track in OGB Large-Scale Challenge\cite{hu2021ogb, ying2021first}. As stated in Section \ref{sec:expressive}, we further find that the proposed Graphormer does not encounter the problem of over-smoothing, i.e., the train and validate error keep going down along with the growth of depth and width of models.

\subsection{Graph Representation}

In this section, we further investigate the performance of Graphormer on commonly used graph-level prediction tasks of popular leaderboards, i.e., OGB~\cite{hu2020open} (OGBG-MolPCBA, OGBG-MolHIV), and benchmarking-GNN~\cite{dwivedi2020benchmarking} (ZINC). Since pre-training is encouraged by OGB, we mainly explore the transferable capability of  a Graphormer model pre-trained on OGB-LSC (i.e., PCQM4M-LSC). Please note that the model configurations, hyper-parameters, and the pre-training performance of pre-trained Graphormers used for MolPCBA and MolHIV are different from the models used in the previous subsection. Please refer to Appendix B for detailed descriptions. For benchmarking-GNN, which does not encourage large pre-trained model, we train an additional Graphormer$_{\small \textsc{Slim}}$\xspace ($L=12, d=80$, total param.$=489K$) from scratch on ZINC.

\paragraph{Baselines.} We report performance of GNNs which achieve top-performance on the official leaderboards\footnote{\url{https://ogb.stanford.edu/docs/leader_graphprop/}\\ \url{https://github.com/graphdeeplearning/benchmarking-gnns/blob/master/docs/07_leaderboards.md}} \emph{without additional domain-specific features}. Considering that the pre-trained Graphormer leverages external data, for a fair comparison on OGB datasets, we additionally report performance for fine-tuning GIN-{\scriptsize VN} pre-trained on PCQM4M-LSC dataset, which achieves the previous state-of-the-art valid and test MAE on that dataset.

\paragraph{Settings.} We report detailed training strategies in Appendix B. 
In addition, Graphormer is more easily trapped in the over-fitting problem due to the large size of the model and the small size of the dataset. Therefore, we employ a widely used data augmentation for graph - FLAG~\cite{kong2020flag}, to mitigate the over-fitting problem on OGB datasets.

\begin{table}[t]
\begin{minipage}{0.47\linewidth}
\centering
\small
\caption{Results on MolPCBA.}
\label{tab:pcba}
\begin{tabular}{c|c|c}
\toprule
  method   & \#param. & AP (\%) \\
  \hline
    DeeperGCN-{\scriptsize VN+FLAG}~\cite{li2020deepergcn} & 5.6M & 28.42$\pm$0.43\\
    DGN~\cite{beaini2020directional} & 6.7M & 28.85$\pm$0.30 \\
    GINE-{\scriptsize VN}~\cite{brossard2020graph} & 6.1M & 29.17$\pm$0.15 \\
    PHC-GNN~\cite{le2021parameterized} & 1.7M & 29.47$\pm$0.26 \\
    GINE-{\scriptsize APPNP}~\cite{brossard2020graph} & 6.1M & 29.79$\pm$0.30 \\
\hline
GIN-{\scriptsize VN}\cite{xu2018how} (fine-tune) &3.4M& 29.02$\pm$0.17 \\
\hline
Graphormer-{\scriptsize FLAG} &119.5M& \textbf{31.39}$\pm$0.32  \\
\bottomrule
\end{tabular}
\end{minipage}
\begin{minipage}{0.65\linewidth}
\centering
\small
\caption{Results on MolHIV.}
\label{tab:hiv}
\begin{tabular}{c|c|c}
\toprule
  method   & \#param. & AUC (\%) \\
  \hline
GCN-{\scriptsize GraphNorm}~\cite{brossard2020graph,cai2020graphnorm} & 526K & 78.83$\pm$1.00 \\
PNA~\cite{corso2020principal} & 326K & 79.05$\pm$1.32 \\
PHC-GNN~\cite{le2021parameterized} & 111K & 79.34$\pm$1.16 \\
DeeperGCN-{\scriptsize FLAG}~\cite{li2020deepergcn} & 532K & 79.42$\pm$1.20\\
DGN~\cite{beaini2020directional} & 114K & 79.70$\pm$0.97 \\
\hline
GIN-{\scriptsize VN}\cite{xu2018how} (fine-tune) &3.3M& 77.80$\pm$1.82 \\
\hline
Graphormer-{\scriptsize FLAG} &47.0M& \textbf{80.51}$\pm$0.53  \\
\bottomrule
\end{tabular}
\end{minipage}
\end{table}

\begin{table}[ht]
\small
\centering
\caption{Results on ZINC. }
\label{tab:zinc}
\begin{tabular}{c|c|c}
\toprule
method                & \#param.  & test MAE \\ \hline
GIN~\cite{xu2018how}  & 509,549   & 0.526$\pm$0.051 \\
GraphSage~\cite{hamilton2017inductive}  & 505,341   & 0.398$\pm$0.002 \\
GAT~\cite{velivckovic2018graph}  & 531,345 & 0.384$\pm$0.007 \\
GCN~\cite{kipf2016semi}  & 505,079   & 0.367$\pm$0.011	 \\
GatedGCN-PE~\cite{bresson2017residual} & 505,011   & 0.214$\pm$0.006 \\
MPNN (sum)~\cite{gilmer2017neural}  & 480,805   & 0.145$\pm$0.007 \\
PNA~\cite{corso2020principal}         & 387,155   & 0.142$\pm$0.010 \\
\hline
GT~\cite{dwivedi2021generalization} & 588,929 & 0.226$\pm$0.014 \\
SAN~\cite{Kreuzer2021rethinking} & 508, 577 & 0.139$\pm$0.006 \\
\hline
Graphormer$_{\small \textsc{SLIM}}$ & 489,321 & \textbf{0.122}$\pm$0.006  \\ 
\bottomrule
\end{tabular}
\end{table}

\begin{table}[ht]
\small
\centering
\caption{
Ablation study results on PCQM4M-LSC dataset with different designs. }
\label{tab:ablation-table}
\begin{tabular}{cccccccc}
\toprule
\multicolumn{2}{c}{Node Relation Encoding} & \multirow{2}{*}{Centrality} & \multicolumn{3}{c}{Edge Encoding}  & \multirow{2}{*}{valid MAE} \\ \cline{1-2} \cline{4-6}
 Laplacian PE\cite{dwivedi2021generalization}&Spatial&&via node&via Aggr&via attn bias(Eq.\ref{eqn:attn-edge})&&\\ \hline
  -&-&-&-&-&-&0.2276\\ \hline
  \yes&-&-&-&-&-&0.1483\\ \hline
  -&\yes&-&-&-&-&0.1427\\ \hline
  -&\yes&\yes&-&-&-&0.1396\\ \hline
  -&\yes&\yes&\yes&-&-&0.1328\\ \hline
  -&\yes&\yes&-&\yes&-&0.1327\\ \hline
  -&\yes&\yes&-&-&\yes&0.1304\\
\bottomrule
\end{tabular}
\end{table}

\paragraph{Results.} Table \ref{tab:pcba}, \ref{tab:hiv} and \ref{tab:zinc} summarize performance of Graphormer comparing with other GNNs on MolHIV, MolPCBA and ZINC datasets. Especially, GT~\cite{dwivedi2021generalization} and SAN~\cite{Kreuzer2021rethinking} in Table \ref{tab:zinc} are recently proposed Transformer-based GNN models. Graphormer consistently and significantly outperforms previous state-of-the-art GNNs on all three datasets by a large margin. Specially, except Graphormer, the other pre-trained GNNs do not achieve competitive performance, which is in line with previous literature~\cite{hu2020strategies}. In addition, we conduct more comparisons to fine-tuning the pre-trained GNNs, please refer to Appendix C.

\subsection{Ablation Studies}
We perform a series of ablation studies on the importance of designs in our proposed Graphormer, on PCQM4M-LSC dataset. The ablation results are included in Table \ref{tab:ablation-table}. To save the computation resources, the Transformer models in table \ref{tab:ablation-table} have 12 layers, and are trained for 100K iterations.

\paragraph{Node Relation Encoding.} We compare previously used positional encoding (PE) to our proposed spatial encoding, which both aim to encode the information of distinct node relation to Transformers. There are various PEs employed by previous Transformer-based GNNs, e.g., Weisfeiler-Lehman-PE (WL-PE)~\cite{zhang2020graph} and Laplacian PE~\cite{belkin2003laplacian,dwivedi2020benchmarking}. We report the performance for Laplacian PE since it performs well comparing to a series of PEs for Graph Transformer in previous literature~\cite{dwivedi2021generalization}. Transformer architecture with the spatial encoding outperforms the counterpart built on the positional encoding, which demonstrates the effectiveness of using spatial encoding to capture the node spatial information.

\paragraph{Centrality Encoding.} Transformer architecture with degree-based centrality encoding yields a large margin performance boost in comparison to those without centrality information. This indicates that the centrality encoding is indispensable to Transformer architecture for modeling graph data.

\paragraph{Edge Encoding.} We compare our proposed edge encoding (denoted as via attn bias) to two commonly used edge encodings described in Section \ref{sec:edge} to incorporate edge features into GNN, denoted as via node and via Aggr in Table \ref{tab:ablation-table}. From the table, the gap of performance is minor between the two conventional methods, but our proposed edge encoding performs significantly better, which indicates that edge encoding as attention bias is more effective for Transformer to capture spatial information on edges.

\section{Related Work}
In this section, we highlight the most recent works which attempt to develop standard Transformer architecture-based GNN or graph structural encoding, but spend less effort on elaborating the works by adapting attention mechanism to GNNs~\cite{li2019graph,yun2019graph,cai2020graph,hu2020heterogeneous,baek2021accurate,velivckovic2018graph,wang2020direct,zhang2020graph,shi2020masked}.

\subsection{Graph Transformer}

There are several works that study the performance of pure Transformer architectures (stacked by transformer layers) with modifications on graph representation tasks, which are more related to our Graphormer. For example,  several parts of the transformer layer are modified in ~\cite{rong2020self}, including an additional GNN employed in attention sub-layer to produce vectors of $Q$, $K$, and $V$, long-range residual connection, and two branches of FFN to produce node and edge representations separately. They pre-train their model on 10 million unlabelled molecules and achieve excellent results by fine-tuning on downstream tasks. Attention module is modified to a soft adjacency matrix in ~\cite{maziarka2020molecule} by directly adding the adjacency matrix and RDKit\footnote{\url{https://www.rdkit.org/}}-computed inter-atomic distance matrix to the attention probabilites. Very recently, Dwivedi \textit{et al.} \cite{dwivedi2021generalization} revisit a series of works for Transformer-based GNNs, and suggest that the attention mechanism in Transformers on graph data should only aggregate the information from neighborhood (i.e., using adjacent matrix as attention mask) to ensure graph sparsity, and propose to use Laplacian eigenvector as positional encoding. Their model GT surpasses baseline GNNs on graph representation task. A concurrent work~\cite{Kreuzer2021rethinking} propose a novel full Laplacian spectrum to learn the position of each node in a graph, and empirically shows better results than GT.

\subsection{Structural Encodings in GNNs}
\paragraph{Path and Distance in GNNs.} Information of path and distance is commonly used in GNNs. For example,  an attention-based aggregation is proposed in ~\cite{chen2019path} where the node features, edge features, one-hot feature of the distance and ring flag feature are concatenated to calculate the attention probabilites; similar to~\cite{chen2019path}, path-based attention is leveraged in~\cite{yangspagan} to model the influence between the center node and its higher-order neighbors; a distance-weighted aggregation scheme on graph is proposed in~\cite{you2019position}; it has been proved in~\cite{li2020distance} that  adopting distance encoding (i.e., one-hot feature of the distance as extra node attribute) could lead to a strictly more expressive power than the 1-WL test.

\paragraph{Positional Encoding in Transformer on Graph.} Several works introduce positional encoding (PE) to Transformer-based GNNs to help the model capture the node position information. For example, Graph-BERT~\cite{zhang2020graph} introduces three types of PE to embed the node position information to model, i.e., an absolute WL-PE which represents different nodes labeled by Weisfeiler-Lehman algorithm, an intimacy based PE and a hop based PE which are both variant to the sampled subgraphs. Absolute Laplacian PE is employed in ~\cite{dwivedi2021generalization} and empircal study shows that its performance surpasses the absolute WL-PE used in ~\cite{zhang2020graph}.

\paragraph{Edge Feature.} Except the conventionally used methods to encode edge feature, which are described in previous section, there are several attempts that exploit how to better encode edge features: an attention-based GNN layer is developed in~\cite{gong2019exploiting} to encode edge features, where the edge feature is weighted by the similarity of the features of its two nodes; edge feature has been encoded into the popular GIN~\cite{xu2018how} in~\cite{brossard2020graph}; in~\cite{dwivedi2021generalization}, the authors propose to project edge features to an embedding vector, then multiply it by attention coefficients, and send the result to an additional FFN sub-layer to produce edge representations;

\section{Conclusion}\label{conclusion}
We have explored the direct application of Transformers to graph representation. With three novel graph structural encodings, the proposed Graphormer works surprisingly well on a wide range of popular benchmark datasets. While these initial results are encouraging, many challenges remain. For example, the quadratic complexity of the self-attention module restricts Graphormer's application on large graphs. Therefore, future development of efficient Graphormer is necessary. Performance improvement could be expected by leveraging domain knowledge-powered encodings on particular graph datasets. Finally, an applicable graph sampling strategy is desired for node representation extraction with Graphormer. We leave them for future works.

\section{Acknowledgement}
We would like to thank Mingqi Yang and Shanda Li for insightful discussions.

\small

\bibliography{ref}
\bibliographystyle{plain}


\newpage
\appendix

\section{Proofs}\label{App:proofs}
\subsection{SPD can Be Used to Improve WL-Test}
\begin{figure}[ht]
    \centering
    \includegraphics[width=0.8\textwidth]{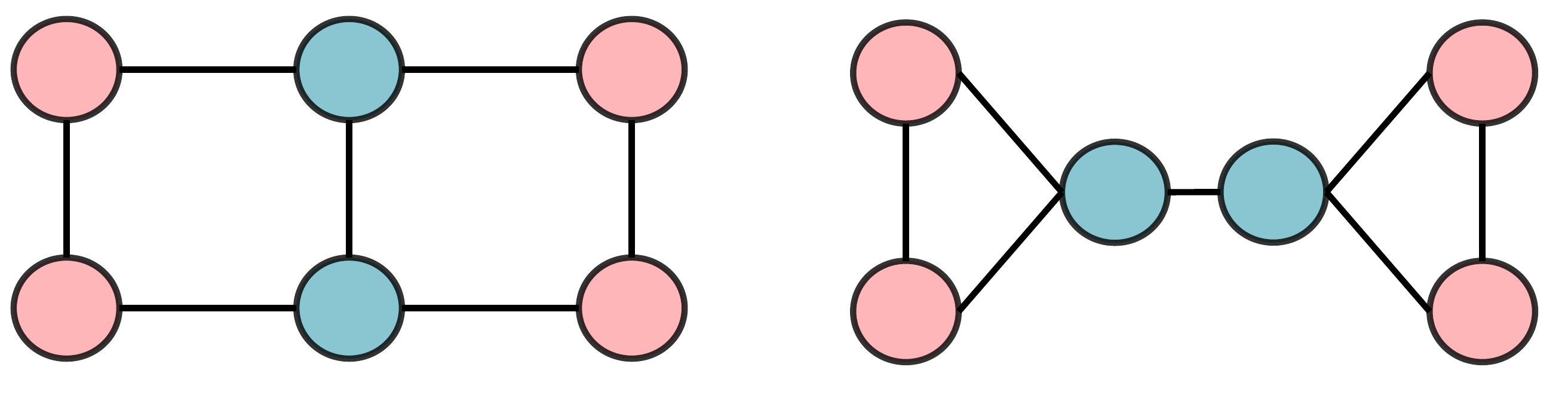}
    \caption{These two graphs cannot be distinguished by 1-WL-test. But the SPD sets, i.e., the SPD from each node to others, are different: The two types of nodes in the left graph have SPD sets $\cbr{0, 1, 1, 2, 2, 3}, \cbr{0, 1, 1, 1, 2, 2}$ while the nodes in the right graph have SPD sets $\cbr{0, 1, 1, 2, 3, 3}, \cbr{0, 1, 1, 1, 2, 2}$.}\label{fig:wl_fail}

\end{figure}

1-WL-test fails in many cases~\cite{NEURIPS2019_bb04af0f,li2020distance}, thus classic message passing GNNs also fail to distinguish many pairs of graphs. We show that SPD might help when 1-WL-test fails, for example, in Figure~\ref{fig:wl_fail} where 1-WL-test fails, the sets of SPD from all nodes to others successfully distinguish the two graphs.

\subsection{Proof of Fact~\ref{fact:recover_gnn}}
\paragraph{MEAN AGGREGATE.} We begin by showing that self-attention module with Spatial Encoding can represent MEAN aggregation. This is achieved by in Eq.~\eqref{eqn:rnpe}: 1) setting $b_\phi = 0$ if $\phi = 1$ and $b_\phi = -\infty$ otherwise where $\phi$ is the SPD; 2) setting $W_Q = W_K = 0$ and $W_V$ to be the identity matrix. Then $\softmax{A}V$ gives the average of representations of the neighbors. 

\paragraph{SUM AGGREGATE.} The SUM aggregation can be realized by first perform MEAN aggregation and then multiply the node degrees. Specifically, the node degrees can be extracted from Centrality Encoding by an additional head and be concatenated to the representations after MEAN aggregation. Then the FFN module in Graphormer can represent the function of multiplying the degree to the dimensions of averaged representations by the universal approximation theorem of FFN.

\paragraph{MAX AGGREGATE.} Representing the MAX aggregation is harder than MEAN and SUM. For each dimension $t$ of the representation vector, we need one head to select the maximal value over $t$-th dimension in the neighbor by in Eq.~\eqref{eqn:rnpe}: 1) setting $b_\phi = 0$ if $\phi = 1$ and $b_\phi = -\infty$ otherwise where $\phi$ is the SPD; 2) setting $W_K = e_t$ which is the $t$-th standard basis; $W_Q = 0$ and the bias term (which is ignored in the previous description for simplicity) of $Q$ to be $T\one$; and $W_V = e_t$, where $T$ is the temperature that can be chosen to be large enough so that the softmax function can approximate hard max and $\one$ is the vector whose elements are all 1.

\paragraph{COMBINE.} The COMBINE step takes the result of AGGREGATE and the previous representation of current node as input. This can be achieved by the AGGREGATE operations described above together with an additional head which outputs the features of present nodes, i.e., in Eq.~\eqref{eqn:rnpe}: 1) setting $b_\phi = 0$ if $\phi = 0$ and $b_\phi = -\infty$ otherwise where $\phi$ is the SPD; 2) setting $W_Q = W_K = 0$ and $W_V$ to be the identity matrix. Then the FFN module can approximate any COMBINE function by the universal approximation theorem of FFN.
\subsection{Proof of Fact~\ref{fact:readout}}
\paragraph{MEAN READOUT.} This can be proved by setting $W_Q = W_K = 0$, the bias terms of $Q, K$ to be $T\one$, and $W_V$ to be the identity matrix where $T$ should be much larger than the scale of $b_\phi$ so that $T^2\one\one^\top$ dominates the Spatial Encoding term.

\section{Experiment Details}

\begin{table}[ht]
\caption{Statistics of the datasets.}
\centering\label{tab:datasets}
\begin{tabular}{cccccccc}
\toprule
Dataset     & Scale & \# Graphs & \# Nodes & \# Edges & Task Type  \\ \hline
PCQM4M-LSC  & Large & 3,803,453 & 53,814,542	& 55,399,880 & Regression  \\ \hline
OGBG-MolPCBA & Medium & 437,929   & 11,386,154& 12,305,805& Binary classification \\ \hline
OGBG-MolHIV & Small & 41,127	& 1,048,738 & 1,130,993 & Binary classification \\ \hline
ZINC (sub-set) & Small & 12,000  & 277,920 & 597,960 & Regression \\
\bottomrule
\end{tabular}
\end{table}

\subsection{Details of Datasets}
We summarize the datasets used in this work in Table \ref{tab:datasets}. PCQM4m-LSC is a quantum chemistry graph-level prediction task in recent OGB Large-Scale Challenge, originally curated under the PubChemQC project ~\cite{nakata2017pubchemqc}. The task of PCQM4M-LSC is to predict DFT(density functional theory)-calculated HOMO-LUMO energy gap of molecules given their 2D molecular graphs, which is one of the most practically-relevant quantum chemical properties of molecule science. PCQM4M-LSC is unprecedentedly large in scale comparing to other labeled graph-level prediction datasets, which contains more than 3.8M graphs. Besides, we conduct experiments on two molecular graph datasets in popular OGB leaderboards, i.e., OGBG-MolPCBA and OGBG-MolHIV. They are two molecular property prediction datasets with different sizes. The pre-trained knowledge of molecular graph on PCQM4M-LSC could be easily leveraged on these two datasets. We adopt official scaffold split on three datasets following ~\cite{hu2021ogb,hu2020open}. In addition, we employ another popular leaderboard, i.e., benchmarking-gnn~\cite{dwivedi2020benchmarking}. We use the ZINC datasets, which is the most popular real-world molecular dataset to predict graph property regression for contrained solubility, an important chemical property for designing generative GNNs for molecules. Different from the scaffold spliting in OGB, uniform sampling is adopted in ZINC for data splitting.

\subsection{Details of Training Strategies}

\subsubsection{PCQM4M-LSC}

\begin{table*}[ht]
\centering 
\caption{Model Configurations and Hyper-parameters of Graphormer on PCQM4M-LSC. } \label{tab:pcq_details}
\begin{threeparttable}
\begin{tabular}{lcc}
\toprule
& Graphormer$_{\small \textsc{Small}}$ & Graphormer \\ \hline
\textbf{\#Layers} & 6 & 12  \\ 
\textbf{Hidden Dimension $d$} & 512 & 768  \\ 
\textbf{FFN Inner-layer Dimension} & 512 & 768  \\ 
\textbf{\#Attention Heads} & 32 & 32  \\ 
\textbf{Hidden Dimension of Each Head} & 16 & 24  \\ 
\textbf{FFN Dropout} & 0.1 & 0.1  \\ 
\textbf{Attention Dropout} & 0.1 & 0.1 \\ 
\textbf{Embedding Dropout} & 0.0 & 0.0 \\ 
\textbf{Max Steps} & 1$M$ & 1$M$ \\
\textbf{Max Epochs} & 300 & 300 \\ 
\textbf{Peak Learning Rate} & 3e-4 & 2e-4 \\ 
\textbf{Batch Size} & 1024 & 1024 \\ 
\textbf{Warm-up Steps} & 60$K$ & 60$K$ \\ 
\textbf{Learning Rate Decay} & Linear & Linear  \\ 
\textbf{Adam $\epsilon$} & 1e-8 & 1e-8 \\ 
\textbf{Adam ($\beta_1$, $\beta_2$)} &  (0.9, 0.999) & (0.9, 0.999) \\ 
\textbf{Gradient Clip Norm} &  5.0 & 5.0  \\ 
\textbf{Weight Decay} & 0.0 & 0.0  \\ 
\bottomrule
\end{tabular}
\end{threeparttable}
\end{table*}

We report the detailed hyper-parameter settings used for training Graphormer in Table \ref{tab:pcq_details}. We reduce the FFN inner-layer dimension of $4d$ in ~\cite{vaswani2017attention} to $d$, which does not appreciably hurt the performance but significantly save the parameters. The embedding dropout ratio is set to 0.1 by default in many previous Transformer works~\cite{devlin2019bert,liu2019roberta}. However, we empirically find that a small embedding dropout ratio (e.g., 0.1) would lead to an observable performance drop on validation set of PCQM4M-LSC. One possible reason is that the molecular graph is relative small (i.e., the median of \#atoms in each molecule is about 15), making graph property more sensitive to the embeddings of each node. Therefore, we set embedding dropout ratio to 0 on this dataset.

\subsubsection{OGBG-MolPCBA}

\begin{table*}[ht]
\centering 
\caption{Hyper-parameters for Graphormer on OGBG-MolPCBA, where the \textbf{text in bold} denotes the hyper-parameters we eventually use. } \label{tab:pcba_details}
\begin{threeparttable}
\begin{tabular}{lc}
\toprule
& Graphormer \\ \hline
\textbf{Max Epochs} & \{2, 5, \textbf{10}\} \\
\textbf{Peak Learning Rate} & \{2e-4, \textbf{3e-4}\}  \\ 
\textbf{Batch Size} & 256 \\ 
\textbf{Warm-up Ratio} & 0.06 \\ 
\textbf{Attention Dropout} & 0.3 \\ 
\textbf{$m$} & \{1, 2,3,\textbf{4}\} \\ 
\textbf{$\alpha$} & 0.001 \\ 
\textbf{$\epsilon$} & 0.001 \\ 
\bottomrule
\end{tabular}
\end{threeparttable}
\end{table*}

\paragraph{Pre-training.} We first report the model configurations and hyper-parameters of the pre-trained Graphormer on PCQM4M-LSC. Empirically, we find that the performance on MolPCBA benefits from the large pre-training model size. Therefore, we train a deep Graphormer with 18 Transformer layers on PCQM4M-LSC. The hidden dimension and FFN inner-layer dimension are set to 1024. We set peak learning rate to 1e-4 for the deep Graphormer. Besides, we enlarge the attention dropout ratio from 0.1 to 0.3 in both pre-training and fine-tuning to prevent the model from over-fitting. The rest of hyper-parameters remain unchanged. The pre-trained Graphormer used for MolPCBA achieves a valid MAE of 0.1253 on PCQM4M-LSC, which is slightly worse than the reports in Table \ref{tab:pcq-table}. 

\paragraph{Fine-tuning.} Table \ref{tab:pcba_details} summarizes the hyper-parameters used for fine-tuning Graphormer on OGBG-MolPCBA. We conduct a grid search for several hyper-parameters to find the optimal configuration. The experimental results are reported by the mean of 10 independent runs with random seeds. We use FLAG~\cite{kong2020flag} with minor modifications for graph data augmentation. In particular, except the step size $\alpha$ and the number of steps $m$, we also employ a projection step in ~\cite{zhu2020freelb} with maximum perturbation $\epsilon$. The performance of Graphormer on MolPCBA is quite robust to the hyper-parameters of FLAG. The rest of hyper-parameters are the same with the pre-training model.

\subsubsection{OGBG-MolHIV}

\begin{table*}[h]
\centering 
\caption{Hyper-parameters for Graphormer on OGBG-MolHIV, where the \textbf{text in bold} denotes the hyper-parameters we eventually use. } \label{tab:hiv_details}
\begin{threeparttable}
\begin{tabular}{lc}
\toprule
& Graphormer \\ \hline
\textbf{Max Epochs} & 8 \\
\textbf{Peak Learning Rate} & 2e-4  \\ 
\textbf{Batch Size} & 128 \\ 
\textbf{Warm-up Ratio} & 0.06 \\ 
\textbf{Dropout} & 0.1 \\ 
\textbf{Attention Dropout} & 0.1 \\ 
\textbf{$m$} & \{1,\textbf{2},3,4\} \\ 
\textbf{$\alpha$} & \{0.001, 0.01, 0.1, \textbf{0.2}\} \\ 
\textbf{$\epsilon$} & \{\textbf{0}, 0.001, 0.01, 0.1\} \\ 
\bottomrule
\end{tabular}
\end{threeparttable}
\end{table*}

\paragraph{Pre-training.} We use the Graphormer reported in Table \ref{tab:pcq-table} as the pre-trained model for OGBG-MolHIV, where the pre-training hyper-parameters are summarized in Table \ref{tab:pcq_details}.

\paragraph{Fine-tuning.} The hyper-parameters for fine-tuning Graphormer on OGBG-MolHIV are presented in Table \ref{tab:hiv_details}. Empirically, we find that the different choices of hyper-parameters of FLAG (i.e., step size $\alpha$, number of steps $m$, and maximum perturbation $\epsilon$) would greatly affect the performance of Graphormer on OGBG-MolHiv. Therefore, we spend more effort to conduct grid search for hyper-parameters of FLAG. We report the best hyper-parameters by the mean of 10 independent runs with random seeds.

\subsubsection{ZINC}

To keep the total parameters of Graphormer less than 500K per the request from benchmarking-GNN leaderboard~\cite{dwivedi2020benchmarking}, we train a slim 12-layer Graphormer with hidden dimension of 80, which is called Graphormer$_{\small \textsc{Slim}}$\xspace in Table \ref{tab:zinc}, and has about 489K learnable parameters. The number of attention heads is set to 8. Table \ref{tab:zinc_details} summarizes the detailed hyper-parameters on ZINC. We train 400K steps on this dataset, and employ a weight decay of 0.01.

\begin{table*}[ht]
\centering 
\caption{Model Configurations and Hyper-parameters on ZINC(sub-set). } \label{tab:zinc_details}
\begin{threeparttable}
\begin{tabular}{lc}
\toprule
& Graphormer$_{\small \textsc{Slim}}$ \\ \hline
\textbf{\#Layers} & 12  \\ 
\textbf{Hidden Dimension} & 80  \\ 
\textbf{FFN Inner-Layer Hidden Dimension} & 80  \\ 
\textbf{\#Attention Heads} & 8  \\ 
\textbf{Hidden Dimension of Each Head} & 10  \\ 
\textbf{FFN Dropout} & 0.1  \\ 
\textbf{Attention Dropout} & 0.1 \\ 
\textbf{Embedding Dropout} & 0.0 \\ 
\textbf{Max Steps} & 400$K$ \\
\textbf{Max Epochs} & 10$K$  \\ 
\textbf{Peak Learning Rate} & 2e-4  \\ 
\textbf{Batch Size} & 256 \\ 
\textbf{Warm-up Steps} & 40$K$ \\ 
\textbf{Learning Rate Decay} & Linear \\ 
\textbf{Adam $\epsilon$} & 1e-8 \\ 
\textbf{Adam ($\beta_1$, $\beta_2$)} &  (0.9, 0.999) \\ 
\textbf{Gradient Clip Norm} &  5.0  \\ 
\textbf{Weight Decay} & 0.01 \\ 
\bottomrule
\end{tabular}
\end{threeparttable}
\end{table*}

\subsection{Details of Hyper-parameters for Baseline Methods}
In this section, we present the details of our re-implementation of the baseline methods.

\subsubsection{PCQM4M-LSC}

The official Github repository of OGB-LSC\footnote{\url{https://github.com/snap-stanford/ogb/tree/master/examples/lsc/pcqm4m}} provides hyper-parameters and codes to reproduce the results on leaderboard. These hyper-parameters work well on almost all popular GNN variants, except the DeeperGCN-{\scriptsize VN}, which results in a training divergence. Therefore, for DeeperGCN-{\scriptsize VN}, we follow the official hyper-parameter setting\footnote{\url{https://github.com/lightaime/deep_gcns_torch/tree/master/examples/ogb/ogbg_mol\#train}} provided by the authors~\cite{li2020deepergcn}. For a fair comparison to Graphormer, we train a 12-layer DeeperGCN. The hidden dimension is set to 600. The batch size is set to 256. The learning rate is set to 1e-3, and a step learning rate scheduler is employed with the decaying step size and the decaying factor $\gamma$ as 30 epochs and 0.25. The model is trained for 100 epochs.

The default dimension of laplacian PE of GT~\cite{dwivedi2021generalization} is set to 8. However, it will cause 2.91\% small molecules (less than 8 atoms) to be filtered out. Therefore, for GT and GT-{\scriptsize Wide}, we set the dimension of laplacian PE to 4, which results in only 0.08\% filtering out. We adopt the default hyper-parameter settings described in ~\cite{dwivedi2021generalization}, except that we decrease the learning rate to 1e-4, which leads to a better convergence on PCQM4M-LSC.

\begin{table*}[t]
\centering 
\caption{Hyper-parameters for fine-tuning GROVER on MolHIV and MolPCBA. } \label{tab:grover_details}
\begin{threeparttable}
\begin{tabular}{lcc}
\toprule
& GROVER & GROVER$_{\small \textsc{LARGE}}$ \\ \hline
\textbf{Dropout} & \{0.1, 0.5\} & \{0.1, 0.5\} \\ 
\textbf{Max Epochs} & \{10, 30, 50\} & \{10, 30\} \\ 
\textbf{Learning Rate} & \{5e-5, 1e-4, 5e-4, 1e-3\} & \{5e-5, 1e-4, 5e-4, 1e-3\} \\ 
\textbf{Batch Size} & \{64, 128\} & \{64, 128\} \\ 
\textbf{Initial Learning Rate} & 1e-7 & 1e-7 \\
\textbf{End Learning Rate} & 1e-9 & 1e-9 \\
\bottomrule
\end{tabular}
\end{threeparttable}
\end{table*}

\begin{table}[t]
\centering
\small
\caption{Comparison to pre-trained Transformer-based GNN on MolHIV. * \ indicates that additional features for molecule are used.}\label{tab:hiv-appendix}
\begin{tabular}{c|c|c}
\toprule
  method   & \#param. & AUC (\%) \\
\hline
Morgan Finger Prints + Random Forest*& 230K & \textbf{80.60}$\pm$0.10 \\
\hline
GROVER*\cite{rong2020self} &48.8M & 79.33$\pm$0.09 \\
GROVER$_{\small \textsc{Large}}$*\cite{rong2020self} &107.7M & 80.32$\pm$0.14 \\
\hline
Graphormer-{\scriptsize FLAG} &47.0M& 80.51$\pm$0.53  \\
\bottomrule
\end{tabular}
\end{table}

\begin{table}[t]
\centering
\small
\caption{Comparison to pre-trained Transformer-based GNN on MolPCBA. * \ indicates that additional features for molecule are used.}\label{tab:pcba-appendix}
\begin{tabular}{c|c|c}
\toprule
  method   & \#param. & AP (\%) \\

\hline
GROVER*\cite{rong2020self} &48.8M & 16.77$\pm$0.36 \\
GROVER$_{\small \textsc{Large}}$*\cite{rong2020self} &107.7M & 13.05$\pm$0.18 \\
\hline
Graphormer-{\scriptsize FLAG} &47.0M& \textbf{31.39}$\pm$0.32  \\
\bottomrule
\end{tabular}
\end{table}

\subsubsection{OGBG-MolPCBA}
To fine-tune the pre-trained GIN-{\scriptsize VN} on MolPCBA, we follow the hyper-parameter settings provided in the original OGB paper~\cite{hu2020open}. To be more concrete, we load the pre-trained checkpoint reported in Table \ref{tab:pcq-table} and fine-tune it on OGBG-MolPCBA dataset. We use the grid search on the hyper-parameters for better fine-tuning performance. In particular, the learning rate is selected from $\{1e-5, 1e-4, 1e-3\}$; the dropout ratio is selected from $\{0.0, 0.1, 0.5\}$; the batch size is selected from $\{32, 64\}$. 

\subsubsection{OGBG-MolHIV}
Similarly, we fine-tune the pre-trained GIN-{\scriptsize VN} on MolHIV by following the hyper-parameter settings provided in the original OGB paper~\cite{hu2020open}. We also conduct the grid search to look for optimal hyper-parameters. The ranges for each hyper-parameter of grid search are the same as the previous subsection.

\section{More Experiments}

As described in the related work, GROVER is a Transformer-based GNN, which has 100 million parameters and pre-trained on 10 million unlabelled molecules using 250 Nvidia V100 GPUs. In this section, we report the fine-tuning scores of GROVER on MolHIV and MolPCBA, and compare with proposed Graphormer.

We download the pre-trained GROVER models from its official Github webpage\footnote{\url{https://github.com/tencent-ailab/grover}}, follow the official instructions\footnote{\url{https://github.com/tencent-ailab/grover/blob/main/README.md\#finetuning-with-existing-data}} and fine-tune the provided pre-trained checkpoints with careful search of hyper-parameters (in Table \ref{tab:grover_details}).  We find that GROVER could achieve competitive performance on MolHIV only if employing additional molecular features, i.e., morgan molecular finger prints and 2D features\footnote{\url{https://github.com/tencent-ailab/grover\# optional-molecular-feature-extraction-1}}. Therefore, we report the scores of GROVER by taking these two additional molecular features. Please note that, from the leaderboard\footnote{\url{https://ogb.stanford.edu/docs/leader_graphprop/}}, we can know such additional molecular features are very effective on MolHIV dataset.

Table \ref{tab:hiv-appendix} and \ref{tab:pcba-appendix} summarize the performance of GROVER and GROVER$_{\small \textsc{Large}}$ comparing with Graphormer on MolHIV and MolPCBA. From the tables, we observe that Graphormer could consistently outperform GROVER even without any additional molecular features.

\section{Discussion \& Future Work}\label{sec:discuss}

\paragraph{Complexity.} Similar to regular Transformer, the attention mechanism in Graphormer scales quadratically with the number of nodes $n$ in the input graph, which may be prohibitively expensive for large $n$ and precludes its usage in settings with limited computational resources. Recently, many solutions have been proposed to address this problem in Transformer~\cite{ke2020rethinking,wang2020linformer,ying2021lazyformer,luo2021stable}. This issue would be greatly benefit from the future development of efficient Graphormer.

\paragraph{Choice of centrality and $\phi$.} In Graphormer, there are multiple choices for the network centrality and the spatial encoding function $\phi(v_i,v_j)$. For example, one can leverage the $L_2$ distance in 3D structure between two atoms in a molecule. In this paper, we mainly evaluate general centrality and distance metric in graph theory, i.e., the degree centrality and the shortest path. Performance improvement could be expected by leveraging domain knowledge powered encodings on particular graph dataset.

\paragraph{Node Representation.} There is a wide range of node representation tasks on graph structured data, such as finance, social network, and temporal prediction. Graphormer could be naturally used for node representation extraction with an applicable graph sampling strategy. We leave it for future work. 

\end{document}

%% file: math_commands.tex
\usepackage{mathtools}
\usepackage{amssymb}
\usepackage{latexsym}
\usepackage{dsfont}
\usepackage{mathrsfs}
\usepackage{amssymb}
\usepackage{amsfonts}
\usepackage{bm}
\usepackage{xspace}
\usepackage{amsthm}

\newcommand{\Ncal}{\mathcal{N}}

\newcommand{\Rbb}{\mathbb{R}}

\ifx\BlackBox\undefined
\newcommand{\BlackBox}{\rule{1.5ex}{1.5ex}}  
\fi

\ifx\QED\undefined
\def\QED{~\rule[-1pt]{5pt}{5pt}\par\medskip}
\fi

\ifx\proof\undefined
\newenvironment{proof}{\par\noindent{\em Proof:\ }}{\hfill\BlackBox\\[.0mm]}
\fi

\ifx\theorem\undefined
\newtheorem{theorem}{Theorem}[section]
\fi

\ifx\example\undefined
\newtheorem{example}{Example}[section]
\fi

\ifx\property\undefined
\newtheorem{property}{Property}
\fi

\ifx\lemma\undefined
\newtheorem{lemma}{Lemma}[section]
\fi

\ifx\proposition\undefined
\newtheorem{proposition}{Proposition}[section]
\fi

\ifx\remark\undefined
\newtheorem{remark}{Remark}
\fi

\ifx\corollary\undefined
\newtheorem{corollary}{Corollary}[section]
\fi

\ifx\definition\undefined
\newtheorem{definition}{Definition}[section]
\fi

\ifx\conjecture\undefined
\newtheorem{conjecture}{Conjecture}[section]
\fi

\ifx\axiom\undefined
\newtheorem{axiom}[theorem]{Axiom}
\fi

\ifx\claim\undefined
\newtheorem{claim}[theorem]{Claim}
\fi

\ifx\assumption\undefined
\newtheorem{assumption}{Assumption}
\fi

\ifx\problem\undefined
\newtheorem{problem}{Problem}[section]
\fi

\ifx\fact\undefined
\newtheorem{fact}{Fact}
\fi

\newcommand{\rbr}[1]{\left(#1\right)}
\newcommand{\sbr}[1]{\left[#1\right]}
\newcommand{\cbr}[1]{\left\{#1\right\}}

\newcommand{\abs}[1]{\left|#1\right|}

\newcommand{\one}{\mathbf{1}}  






%% file: main.bbl
\begin{thebibliography}{10}

\bibitem{baek2021accurate}
Jinheon Baek, Minki Kang, and Sung~Ju Hwang.
\newblock Accurate learning of graph representations with graph multiset
  pooling.
\newblock {\em ICLR}, 2021.

\bibitem{beaini2020directional}
Dominique Beaini, Saro Passaro, Vincent L{\'e}tourneau, William~L Hamilton,
  Gabriele Corso, and Pietro Li{\`o}.
\newblock Directional graph networks.
\newblock In {\em International Conference on Machine Learning}, 2021.

\bibitem{belkin2003laplacian}
Mikhail Belkin and Partha Niyogi.
\newblock Laplacian eigenmaps for dimensionality reduction and data
  representation.
\newblock {\em Neural computation}, 15(6):1373--1396, 2003.

\bibitem{bresson2017residual}
Xavier Bresson and Thomas Laurent.
\newblock Residual gated graph convnets.
\newblock {\em arXiv preprint arXiv:1711.07553}, 2017.

\bibitem{brossard2020graph}
R{\'e}my Brossard, Oriel Frigo, and David Dehaene.
\newblock Graph convolutions that can finally model local structure.
\newblock {\em arXiv preprint arXiv:2011.15069}, 2020.

\bibitem{brown2020language}
Tom Brown, Benjamin Mann, Nick Ryder, Melanie Subbiah, Jared~D Kaplan, Prafulla
  Dhariwal, Arvind Neelakantan, Pranav Shyam, Girish Sastry, Amanda Askell,
  Sandhini Agarwal, Ariel Herbert-Voss, Gretchen Krueger, Tom Henighan, Rewon
  Child, Aditya Ramesh, Daniel Ziegler, Jeffrey Wu, Clemens Winter, Chris
  Hesse, Mark Chen, Eric Sigler, Mateusz Litwin, Scott Gray, Benjamin Chess,
  Jack Clark, Christopher Berner, Sam McCandlish, Alec Radford, Ilya Sutskever,
  and Dario Amodei.
\newblock Language models are few-shot learners.
\newblock In H.~Larochelle, M.~Ranzato, R.~Hadsell, M.~F. Balcan, and H.~Lin,
  editors, {\em Advances in Neural Information Processing Systems}, volume~33,
  pages 1877--1901. Curran Associates, Inc., 2020.

\bibitem{cai2020graph}
Deng Cai and Wai Lam.
\newblock Graph transformer for graph-to-sequence learning.
\newblock In {\em Proceedings of the AAAI Conference on Artificial
  Intelligence}, volume~34, pages 7464--7471, 2020.

\bibitem{cai2020graphnorm}
Tianle Cai, Shengjie Luo, Keyulu Xu, Di~He, Tie-yan Liu, and Liwei Wang.
\newblock Graphnorm: A principled approach to accelerating graph neural network
  training.
\newblock In {\em International Conference on Machine Learning}, 2021.

\bibitem{chen2019path}
Benson Chen, Regina Barzilay, and Tommi Jaakkola.
\newblock Path-augmented graph transformer network.
\newblock {\em arXiv preprint arXiv:1905.12712}, 2019.

\bibitem{corso2020principal}
Gabriele Corso, Luca Cavalleri, Dominique Beaini, Pietro Li{\`o}, and Petar
  Veli{\v{c}}kovi{\'c}.
\newblock Principal neighbourhood aggregation for graph nets.
\newblock {\em Advances in Neural Information Processing Systems}, 33, 2020.

\bibitem{devlin2019bert}
Jacob Devlin, Ming-Wei Chang, Kenton Lee, and Kristina Toutanova.
\newblock Bert: Pre-training of deep bidirectional transformers for language
  understanding.
\newblock In {\em Proceedings of the 2019 Conference of the North American
  Chapter of the Association for Computational Linguistics: Human Language
  Technologies, Volume 1 (Long and Short Papers)}, pages 4171--4186, 2019.

\bibitem{dosovitskiy2020image}
Alexey Dosovitskiy, Lucas Beyer, Alexander Kolesnikov, Dirk Weissenborn,
  Xiaohua Zhai, Thomas Unterthiner, Mostafa Dehghani, Matthias Minderer, Georg
  Heigold, Sylvain Gelly, et~al.
\newblock An image is worth 16x16 words: Transformers for image recognition at
  scale.
\newblock {\em arXiv preprint arXiv:2010.11929}, 2020.

\bibitem{dwivedi2021generalization}
Vijay~Prakash Dwivedi and Xavier Bresson.
\newblock A generalization of transformer networks to graphs.
\newblock {\em AAAI Workshop on Deep Learning on Graphs: Methods and
  Applications}, 2021.

\bibitem{dwivedi2020benchmarking}
Vijay~Prakash Dwivedi, Chaitanya~K Joshi, Thomas Laurent, Yoshua Bengio, and
  Xavier Bresson.
\newblock Benchmarking graph neural networks.
\newblock {\em arXiv preprint arXiv:2003.00982}, 2020.

\bibitem{gilmer2017neural}
Justin Gilmer, Samuel~S Schoenholz, Patrick~F Riley, Oriol Vinyals, and
  George~E Dahl.
\newblock Neural message passing for quantum chemistry.
\newblock In {\em International Conference on Machine Learning}, pages
  1263--1272. PMLR, 2017.

\bibitem{gong2019exploiting}
Liyu Gong and Qiang Cheng.
\newblock Exploiting edge features for graph neural networks.
\newblock In {\em Proceedings of the IEEE/CVF Conference on Computer Vision and
  Pattern Recognition}, pages 9211--9219, 2019.

\bibitem{gulati2020conformer}
Anmol Gulati, James Qin, Chung-Cheng Chiu, Niki Parmar, Yu~Zhang, Jiahui Yu,
  Wei Han, Shibo Wang, Zhengdong Zhang, Yonghui Wu, et~al.
\newblock Conformer: Convolution-augmented transformer for speech recognition.
\newblock {\em arXiv preprint arXiv:2005.08100}, 2020.

\bibitem{hamilton2017inductive}
William~L Hamilton, Zhitao Ying, and Jure Leskovec.
\newblock Inductive representation learning on large graphs.
\newblock In {\em NIPS}, 2017.

\bibitem{hellendoorn2019global}
Vincent~J Hellendoorn, Charles Sutton, Rishabh Singh, Petros Maniatis, and
  David Bieber.
\newblock Global relational models of source code.
\newblock In {\em International conference on learning representations}, 2019.

\bibitem{hu2020strategies}
W~Hu, B~Liu, J~Gomes, M~Zitnik, P~Liang, V~Pande, and J~Leskovec.
\newblock Strategies for pre-training graph neural networks.
\newblock In {\em International Conference on Learning Representations (ICLR)},
  2020.

\bibitem{hu2021ogb}
Weihua Hu, Matthias Fey, Hongyu Ren, Maho Nakata, Yuxiao Dong, and Jure
  Leskovec.
\newblock Ogb-lsc: A large-scale challenge for machine learning on graphs.
\newblock {\em arXiv preprint arXiv:2103.09430}, 2021.

\bibitem{hu2020open}
Weihua Hu, Matthias Fey, Marinka Zitnik, Yuxiao Dong, Hongyu Ren, Bowen Liu,
  Michele Catasta, and Jure Leskovec.
\newblock Open graph benchmark: Datasets for machine learning on graphs.
\newblock {\em arXiv preprint arXiv:2005.00687}, 2020.

\bibitem{hu2020heterogeneous}
Ziniu Hu, Yuxiao Dong, Kuansan Wang, and Yizhou Sun.
\newblock Heterogeneous graph transformer.
\newblock In {\em Proceedings of The Web Conference 2020}, pages 2704--2710,
  2020.

\bibitem{ishiguro2019graph}
Katsuhiko Ishiguro, Shin-ichi Maeda, and Masanori Koyama.
\newblock Graph warp module: an auxiliary module for boosting the power of
  graph neural networks in molecular graph analysis.
\newblock {\em arXiv preprint arXiv:1902.01020}, 2019.

\bibitem{ke2020rethinking}
Guolin Ke, Di~He, and Tie-Yan Liu.
\newblock Rethinking the positional encoding in language pre-training.
\newblock {\em ICLR}, 2020.

\bibitem{kipf2016semi}
Thomas~N Kipf and Max Welling.
\newblock Semi-supervised classification with graph convolutional networks.
\newblock {\em arXiv preprint arXiv:1609.02907}, 2016.

\bibitem{kong2020flag}
Kezhi Kong, Guohao Li, Mucong Ding, Zuxuan Wu, Chen Zhu, Bernard Ghanem, Gavin
  Taylor, and Tom Goldstein.
\newblock Flag: Adversarial data augmentation for graph neural networks.
\newblock {\em arXiv preprint arXiv:2010.09891}, 2020.

\bibitem{Kreuzer2021rethinking}
Devin Kreuzer, Dominique Beaini, William Hamilton, Vincent Létourneau, and
  Prudencio Tossou.
\newblock Rethinking graph transformers with spectral attention.
\newblock {\em arXiv preprint arXiv:2106.03893}, 2021.

\bibitem{le2021parameterized}
Tuan Le, Marco Bertolini, Frank No{\'e}, and Djork-Arn{\'e} Clevert.
\newblock Parameterized hypercomplex graph neural networks for graph
  classification.
\newblock {\em arXiv preprint arXiv:2103.16584}, 2021.

\bibitem{li2020deepergcn}
Guohao Li, Chenxin Xiong, Ali Thabet, and Bernard Ghanem.
\newblock Deepergcn: All you need to train deeper gcns.
\newblock {\em arXiv preprint arXiv:2006.07739}, 2020.

\bibitem{li2017learning}
Junying Li, Deng Cai, and Xiaofei He.
\newblock Learning graph-level representation for drug discovery.
\newblock {\em arXiv preprint arXiv:1709.03741}, 2017.

\bibitem{li2020distance}
Pan Li, Yanbang Wang, Hongwei Wang, and Jure Leskovec.
\newblock Distance encoding: Design provably more powerful neural networks for
  graph representation learning.
\newblock {\em Advances in Neural Information Processing Systems}, 33, 2020.

\bibitem{li2019graph}
Yuan Li, Xiaodan Liang, Zhiting Hu, Yinbo Chen, and Eric~P. Xing.
\newblock Graph transformer, 2019.

\bibitem{lin2018multi}
Xi~Victoria Lin, Richard Socher, and Caiming Xiong.
\newblock Multi-hop knowledge graph reasoning with reward shaping.
\newblock {\em arXiv preprint arXiv:1808.10568}, 2018.

\bibitem{liu2019roberta}
Yinhan Liu, Myle Ott, Naman Goyal, Jingfei Du, Mandar Joshi, Danqi Chen, Omer
  Levy, Mike Lewis, Luke Zettlemoyer, and Veselin Stoyanov.
\newblock Roberta: A robustly optimized bert pretraining approach.
\newblock {\em arXiv preprint arXiv:1907.11692}, 2019.

\bibitem{liu2021Swin}
Ze~Liu, Yutong Lin, Yue Cao, Han Hu, Yixuan Wei, Zheng Zhang, Stephen Lin, and
  Baining Guo.
\newblock Swin transformer: Hierarchical vision transformer using shifted
  windows.
\newblock {\em arXiv preprint arXiv:2103.14030}, 2021.

\bibitem{luo2021stable}
Shengjie Luo, Shanda Li, Tianle Cai, Di~He, Dinglan Peng, Shuxin Zheng, Guolin
  Ke, Liwei Wang, and Tie-Yan Liu.
\newblock Stable, fast and accurate: Kernelized attention with relative
  positional encoding.
\newblock {\em NeurIPS}, 2021.

\bibitem{NEURIPS2019_bb04af0f}
Haggai Maron, Heli Ben-Hamu, Hadar Serviansky, and Yaron Lipman.
\newblock Provably powerful graph networks.
\newblock In H.~Wallach, H.~Larochelle, A.~Beygelzimer, F.~d\textquotesingle
  Alch\'{e}-Buc, E.~Fox, and R.~Garnett, editors, {\em Advances in Neural
  Information Processing Systems}, volume~32. Curran Associates, Inc., 2019.

\bibitem{marshall2010promotion}
P~David Marshall.
\newblock The promotion and presentation of the self: celebrity as marker of
  presentational media.
\newblock {\em Celebrity studies}, 1(1):35--48, 2010.

\bibitem{marwick2011see}
Alice Marwick and Danah Boyd.
\newblock To see and be seen: Celebrity practice on twitter.
\newblock {\em Convergence}, 17(2):139--158, 2011.

\bibitem{maziarka2020molecule}
{\L}ukasz Maziarka, Tomasz Danel, S{\l}awomir Mucha, Krzysztof Rataj, Jacek
  Tabor, and Stanis{\l}aw Jastrz{\k{e}}bski.
\newblock Molecule attention transformer.
\newblock {\em arXiv preprint arXiv:2002.08264}, 2020.

\bibitem{nakata2017pubchemqc}
Maho Nakata and Tomomi Shimazaki.
\newblock Pubchemqc project: a large-scale first-principles electronic
  structure database for data-driven chemistry.
\newblock {\em Journal of chemical information and modeling}, 57(6):1300--1308,
  2017.

\bibitem{narang2021transformer}
Sharan Narang, Hyung~Won Chung, Yi~Tay, William Fedus, Thibault Fevry, Michael
  Matena, Karishma Malkan, Noah Fiedel, Noam Shazeer, Zhenzhong Lan, et~al.
\newblock Do transformer modifications transfer across implementations and
  applications?
\newblock {\em arXiv preprint arXiv:2102.11972}, 2021.

\bibitem{peng2021could}
Dinglan Peng, Shuxin Zheng, Yatao Li, Guolin Ke, Di~He, and Tie-Yan Liu.
\newblock How could neural networks understand programs?
\newblock In {\em International Conference on Machine Learning}. PMLR, 2021.

\bibitem{raffel2019exploring}
Colin Raffel, Noam Shazeer, Adam Roberts, Katherine Lee, Sharan Narang, Michael
  Matena, Yanqi Zhou, Wei Li, and Peter~J. Liu.
\newblock Exploring the limits of transfer learning with a unified text-to-text
  transformer.
\newblock {\em Journal of Machine Learning Research}, 21(140):1--67, 2020.

\bibitem{rong2020self}
Yu~Rong, Yatao Bian, Tingyang Xu, Weiyang Xie, Ying Wei, Wenbing Huang, and
  Junzhou Huang.
\newblock Self-supervised graph transformer on large-scale molecular data.
\newblock {\em Advances in Neural Information Processing Systems}, 33, 2020.

\bibitem{shaw2018self}
Peter Shaw, Jakob Uszkoreit, and Ashish Vaswani.
\newblock Self-attention with relative position representations.
\newblock In {\em Proceedings of the 2018 Conference of the North American
  Chapter of the Association for Computational Linguistics: Human Language
  Technologies, Volume 2 (Short Papers)}, pages 464--468, 2018.

\bibitem{shi2020masked}
Yunsheng Shi, Zhengjie Huang, Wenjin Wang, Hui Zhong, Shikun Feng, and Yu~Sun.
\newblock Masked label prediction: Unified message passing model for
  semi-supervised classification.
\newblock {\em arXiv preprint arXiv:2009.03509}, 2020.

\bibitem{vaswani2017attention}
Ashish Vaswani, Noam Shazeer, Niki Parmar, Jakob Uszkoreit, Llion Jones,
  Aidan~N Gomez, Lukasz Kaiser, and Illia Polosukhin.
\newblock Attention is all you need.
\newblock In {\em NIPS}, 2017.

\bibitem{velivckovic2018graph}
Petar Veli{\v{c}}kovi{\'c}, Guillem Cucurull, Arantxa Casanova, Adriana Romero,
  Pietro Lio, and Yoshua Bengio.
\newblock Graph attention networks.
\newblock {\em ICLR}, 2018.

\bibitem{wang2020direct}
Guangtao Wang, Rex Ying, Jing Huang, and Jure Leskovec.
\newblock Direct multi-hop attention based graph neural network.
\newblock {\em arXiv preprint arXiv:2009.14332}, 2020.

\bibitem{wang2020linformer}
Sinong Wang, Belinda Li, Madian Khabsa, Han Fang, and Hao Ma.
\newblock Linformer: Self-attention with linear complexity.
\newblock {\em arXiv preprint arXiv:2006.04768}, 2020.

\bibitem{xiong2020layer}
Ruibin Xiong, Yunchang Yang, Di~He, Kai Zheng, Shuxin Zheng, Chen Xing,
  Huishuai Zhang, Yanyan Lan, Liwei Wang, and Tieyan Liu.
\newblock On layer normalization in the transformer architecture.
\newblock In {\em International Conference on Machine Learning}, pages
  10524--10533. PMLR, 2020.

\bibitem{xu2018how}
Keyulu Xu, Weihua Hu, Jure Leskovec, and Stefanie Jegelka.
\newblock How powerful are graph neural networks?
\newblock In {\em International Conference on Learning Representations}, 2019.

\bibitem{yang2020breaking}
Mingqi Yang, Yanming Shen, Heng Qi, and Baocai Yin.
\newblock Breaking the expressive bottlenecks of graph neural networks.
\newblock {\em arXiv preprint arXiv:2012.07219}, 2020.

\bibitem{yangspagan}
Yiding Yang, Xinchao Wang, Mingli Song, Junsong Yuan, and Dacheng Tao.
\newblock Spagan: Shortest path graph attention network.
\newblock {\em Advances in IJCAI}, 2019.

\bibitem{ying2021lazyformer}
Chengxuan Ying, Guolin Ke, Di~He, and Tie-Yan Liu.
\newblock Lazyformer: Self attention with lazy update.
\newblock {\em arXiv preprint arXiv:2102.12702}, 2021.

\bibitem{ying2021first}
Chengxuan Ying, Mingqi Yang, Shuxin Zheng, Guolin Ke, Shengjie Luo, Tianle Cai,
  Chenglin Wu, Yuxin Wang, Yanming Shen, and Di~He.
\newblock First place solution of kdd cup 2021 \& ogb large-scale challenge
  graph-level track.
\newblock {\em arXiv preprint arXiv:2106.08279}, 2021.

\bibitem{you2019position}
Jiaxuan You, Rex Ying, and Jure Leskovec.
\newblock Position-aware graph neural networks.
\newblock In {\em International Conference on Machine Learning}, pages
  7134--7143. PMLR, 2019.

\bibitem{yun2019graph}
Seongjun Yun, Minbyul Jeong, Raehyun Kim, Jaewoo Kang, and Hyunwoo~J Kim.
\newblock Graph transformer networks.
\newblock {\em Advances in Neural Information Processing Systems}, 32, 2019.

\bibitem{zhang2020graph}
Jiawei Zhang, Haopeng Zhang, Congying Xia, and Li~Sun.
\newblock Graph-bert: Only attention is needed for learning graph
  representations.
\newblock {\em arXiv preprint arXiv:2001.05140}, 2020.

\bibitem{zhu2020freelb}
Chen Zhu, Yu~Cheng, Zhe Gan, Siqi Sun, Tom Goldstein, and Jingjing Liu.
\newblock Freelb: Enhanced adversarial training for natural language
  understanding.
\newblock In {\em ICLR}, 2020.

\bibitem{zugner2020language}
Daniel Z{\"u}gner, Tobias Kirschstein, Michele Catasta, Jure Leskovec, and
  Stephan G{\"u}nnemann.
\newblock Language-agnostic representation learning of source code from
  structure and context.
\newblock In {\em International Conference on Learning Representations}, 2020.

\end{thebibliography}
